\documentclass[10pt,twocolumn,letterpaper]{article}
\usepackage{cvpr}
\usepackage{times}
\usepackage{epsfig}
\usepackage{graphicx}
\usepackage{amsmath}
\usepackage{amssymb}
\usepackage{floatrow}
\usepackage{boldline,multirow}
\usepackage[pagebackref=true,breaklinks=true,letterpaper=true,colorlinks,bookmarks=false]{hyperref}
\usepackage{subcaption}
\usepackage{comment}
\cvprfinalcopy
\usepackage{nopageno}

\DeclareMathOperator*{\argmax}{argmax}

\def\cvprPaperID{4534} 

\ifcvprfinal\fi
\newcommand{\norm}[1]{\left\lVert#1\right\rVert}

\begin{document}

\title{Robust Object Detection under Occlusion with \\Context-Aware CompositionalNets}

\author{
	Angtian Wang\thanks{Joint first authors}
	\;\; Yihong Sun\footnotemark[1]
	\;\; Adam Kortylewski\thanks{Joint senior authors}
	\;\; Alan Yuille\footnotemark[2]\\	
	Johns Hopkins University\\	
}

\maketitle

\begin{abstract}
Detecting partially occluded objects is a difficult task.
Our experimental results show that deep learning approaches, such as Faster R-CNN, are not robust at object detection under occlusion.
Compositional convolutional neural networks (CompositionalNets) have been shown to be robust at classifying occluded objects by explicitly representing the object as a composition of parts.
In this work, we propose to overcome two limitations of CompositionalNets which will enable them to detect partially occluded objects:
1) CompositionalNets, as well as other DCNN architectures, do not explicitly separate the representation of the context from the object itself. 
Under strong object occlusion, the influence of the context is amplified which can have severe negative effects for detection at test time.
In order to overcome this, we propose to segment the context during training via bounding box annotations.
We then use the segmentation to learn a context-aware CompositionalNet that disentangles the representation of the context and the object. 
2) We extend the part-based voting scheme in CompositionalNets to vote for the corners of the object's bounding box, which enables the model to reliably estimate bounding boxes for partially occluded objects.
Our extensive experiments show that our proposed model can detect objects robustly, 
increasing the detection performance of strongly occluded vehicles from PASCAL3D+ and MS-COCO by 41\% and 35\% respectively in absolute performance relative to Faster R-CNN.
\end{abstract}
\section{Introduction}

\begin{figure}
    \centering
    \includegraphics[height=4.5cm]{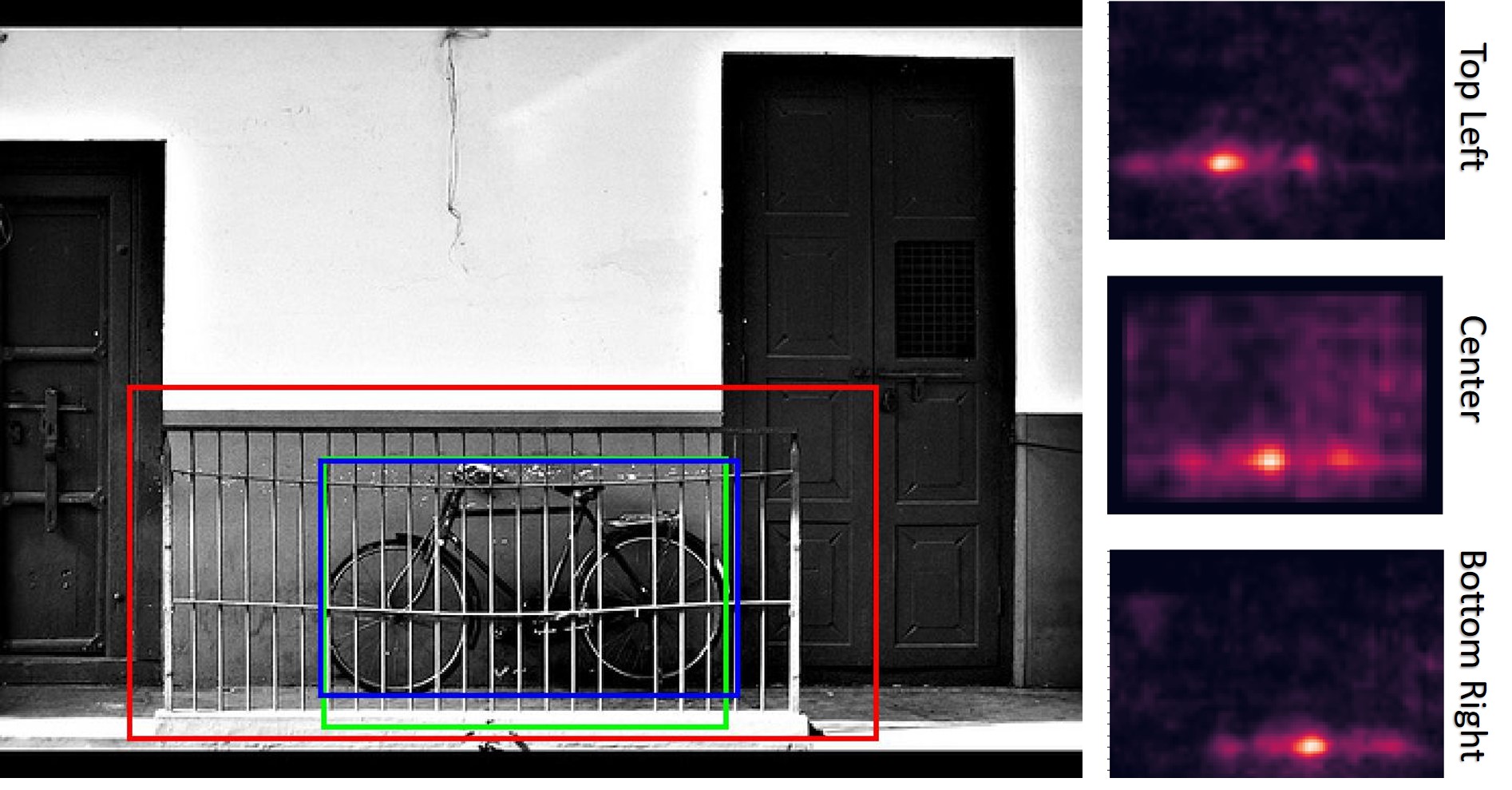}
    \caption{
    Bicycle detection result for an image of the MS-COCO dataset. Blue box: ground truth; red box: detection result by Faster R-CNN; green box: detection result by context-aware CompositionalNet. 
    Probability maps of three-point detection are to the right.
    The proposed context-aware CompositionalNet are able to detect the partially occluded object robustly.
    }
    \label{fig:inodoro}
\end{figure}

In natural images, objects are surrounded and partially occluded by other objects. 
Recognizing partially occluded objects is a difficult task since the appearances and shapes of occluders are highly variable.
Recent work \cite{hongru,kortylewski2019compositional} has shown that deep learning approaches are significantly less robust than humans at classifying partially occluded objects.
Our experimental results show that this limitation of deep learning approaches is even amplified in object detection.
In particular, we find that Faster R-CNN is not robust under partial occlusion, even when it is trained with strong data augmentation with partial occlusion.
Our experiments show that this is caused by two factors: 
1) The proposal network does not localize objects accurately under strong occlusion.
2) The classification network does not classify partially occluded objects robustly. 
Thus, our work highlights key limitations of deep learning approaches to object detection under partial occlusion that need to be addressed.

In contrast to deep convolutional neural networks (DCNNs), compositional models can robustly classify partially occluded objects from a fixed viewpoint \cite{george2017generative,kortylewski2017model} and detect semantic parts of partially occluded object \cite{wang2017detecting,zhang2018deepvoting}.
These models are inspired by the compositionality of human cognition \cite{bienenstock1998compositionality,von1987synaptic,fodor1988connectionism,bienenstock1997compositionality} and share similar characteristics with biological vision systems, such as bottom-up sparse compositional encoding and top-down attentional modulations found in the ventral stream \cite{connor1,connor2,connor3}.
Recent work \cite{cvpr20} proposed the Compositional Convolutional Neural Network (CompositionalNet), a generative compositional model of neural feature activations that can robustly classify images of partially occluded objects. 
This model explicitly represents objects as compositions of parts, which are combined with a voting scheme that enables a robust classification based on the spatial configuration of a few visible parts.
However, we find that CompositionalNets as proposed in \cite{cvpr20} are not suitable for object detection because of two major limitations: 
1) CompositionalNets, as well as other DCNN architectures, do not explicitly disentangle the representation of the context from that of the object. 
Our experiments show that this has negative effects on the detection performance since context is often biased in the training data (e.g. airplanes are often found in blue background).
If objects are strongly occluded, the detection thresholds must be lowered. 
This in turn increases the influence of the objects' context and leads to false-positive detections in regions with no object (e.g. if a strongly occluded car must be detected, a false airplane might be detected in the sky, seen in Figure \ref{fig:context_bad}).
2) CompositionalNets lack mechanisms for robustly estimating the bounding box of the object.
Furthermore, our experiments show that region proposal networks do not estimate the bounding boxes robustly when objects are partially occluded. 

In this work, we propose to build on and significantly extend CompositionalNets in order to enable them to detect partially occluded objects robustly. 
In particular, we introduce a detection layer and propose to decompose the image representation as a mixture of context and object representation. 
We obtain such decomposition by generalizing contextual features in the training data via bounding box annotations.
This context-aware image representation enables us to control the influence of the context on the detection result.
Furthermore, we introduce a robust voting mechanism to estimate the bounding box of the object. 
In particular, we extend the part-based voting scheme in CompositionalNets to also vote for two opposite corners of the bounding box in addition to the object center. 

Our extensive experiments show that the proposed context-aware CompositionalNets with robust bounding box estimation detect objects robustly even under severe occlusion (Figure \ref{fig:inodoro}), increasing the detection performance on strongly occluded vehicles from PASCAL3D+ \cite{PASCAL3D_source} and MS-COCO \cite{lin2014microsoft} by 41\% and 35\% respectively in absolute performance relative to Faster R-CNN.
In summary, we make several important contributions in this work:
\begin{enumerate}
    \item We propose to decompose the image representation in CompositionalNets as a mixture model of context and object representation. 
    We demonstrate that such \textbf{context-aware CompositionalNets} allow for precise control of the influence of the object's context on the detection result, hence, increasing the robustness when classifying strongly occluded objects.
    \item We propose a robust \textbf{part-based voting mechanism for bounding box estimation} that enables the accurate estimation of an object's bounding box even under severe occlusion.
    \item Our experiments demonstrate that context-aware CompositionalNets combined with a part-based bounding box estimation \textbf{outperform Faster R-CNN networks at object detection under partial occlusion} by a significant margin.
\end{enumerate}
\section{Related Work}


{\bf Region selection under occlusion.}
The detection of an object involves the estimation of its location, class and bounding box. While a search over the image can be implemented efficiently, e.g. using a scanning window \cite{slidingwindow}, the number of potential bounding boxes is combinatorial with the number of pixels.
The most widely applied approach for solving this problem is to use Region Proposal Networks (RPNs) \cite{girshick2014rich} which enable the learning of fast approaches to object detection \cite{fastrcnn,faster-rcnn, cascade-rcnn}. 
However, our experiments demonstrate that RPNs do not estimate the bounding box of an object correctly under occlusion.

\textbf{Image classification under occlusion.}
The classification network in deep object detection approaches is typically chosen to be a DCNN, such as ResNet \cite{he2016deep} or VGG \cite{simonyan2014very}.
However, recent work \cite{hongru,kortylewski2019compositional} has shown that standard DCNNs are significantly less robust to partial occlusion compared to humans.
A potential approach to overcome this limitation of DCNNs is to use data augmentation with partial occlusion \cite{cutout,yun2019cutmix} or top-down cues \cite{xiao2019tdapnet}. 
However, our experiments demonstrate that data augmentation approaches have only a limited impact on the generalization of DCNNs under occlusion.
In contrast to deep learning approaches, generative compositional models \cite{jin2006context,zhu2008,fidler2014,dai2014unsupervised,kortylewski2017greedy} have proven to be robust to partial occlusion in the context of detecting object parts \cite{wang2017detecting,kortylewski2017model,zhang2018deepvoting} and recognizing objects from a fixed viewpoint \cite{george2017generative,kortylewski2016probabilistic}. 
Additionally, CompositionalNets \cite{cvpr20}, which integrate compositional models with DCNN architecture, were shown to be significantly more robust for image classification under occlusion.

\textbf{Object Detection under occlusion.} 
Sheng \cite{InferenOccDetect} \textit{et al.} propose a boosted cascade framework for detecting partially visible objects. 
However, their approach uses handcrafted features and can only be applied to images where objects are artificially occluded by cutting out image patches.
Additionally, a number of deep learning approaches have been proposed for detecting occluded objects \cite{OR-CNN, OccNet}; however, these methods require detailed part-level annotations to reconstruct the occluded objects.  
Xiang and Savarese \cite{3DAspectlets} propose to use 3D models and to treat occlusion as a multi-label classification task. However, in a real-world scenario, the classes of occluders can be difficult to model in 3D and are often not known a priori (e.g. the particular type of fence in Figure \ref{fig:inodoro}). 
Also, other approaches are based on videos or stereo images \cite{SymmNet, Symmetricstereo}, however, we focus on object detection in still images. 
Most related to our work is part-based voting approaches \cite{deepvoting, occdata} that have proven to work reliably for semantic part detection under occlusion. 
However, these methods assume a fixed size bounding box which limits their applicability in the context of object detection.

In this work, we extend CompositionalNets to context-aware object detectors with a part-based voting mechanism that can robustly estimate the object's bounding box even under very strong partial occlusion.

\section{Object Detection with CompositionalNets}

In Section \ref{sec:vmf} we discuss prior work on CompositionalNets. We propose a generalization of CompositionalNets to detection in Section \ref{sec:detect}, introducing a detection layer and a robust bounding box estimation mechanism.
Finally, we introduce context-aware CompositionalNets in Section \ref{sec:context}, enabling the model to separate the context from the object representation, making it robust to contextual biases in the training data, while still being able to leverage contextual information under strong occlusion.

\textbf{Notation.} The output of a layer $l$ in a DCNN is referenced as \textit{feature map} $F^l = \psi(I,\Omega) \in 
\mathbb{R}^{H \times W \times D}$, where $I$ is the input image and $\Omega$ are the parameters of the feature extractor. 
\textit{Feature vectors} are vectors in the feature map $f^l_p \in \mathbb{R}^D$ at position $p$, where $p$ is defined on the 2D lattice of $F^l$ with $D$ being the number of channels in the layer. 
We omit subscript $l$ in the following for convenience because this layer is fixed a priori in our experiments.

\begin{figure}
    \centering
    \includegraphics[height=5.2cm]{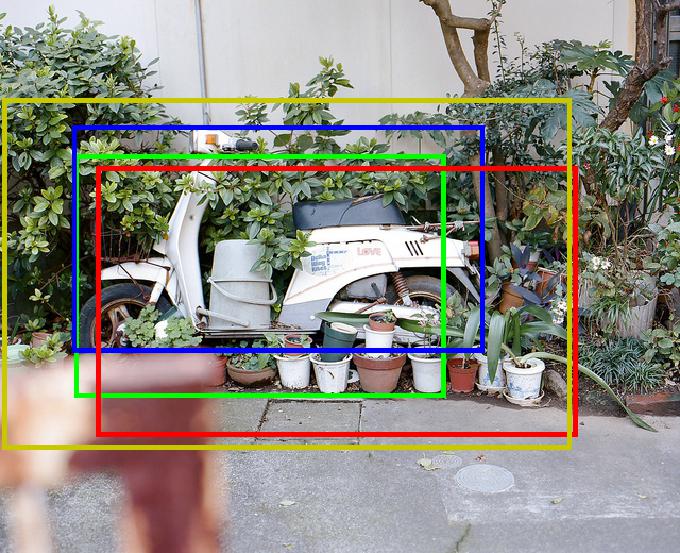}
    \caption{Object detection under occlusion with RPNs and proposed robust bounding box voting. Blue box: ground truth; red box: Faster R-CNN (RPN+VGG); yellow box: RPN+CompositionalNet; green box: context-aware CompositionalNet with robust bounding box voting. Note how the RPN-based approaches fail to localize the object, while our proposed approach can accurately localize the object. 
    }
    \label{fig:rpn}
\end{figure}

\subsection{Prior work: CompositionalNets}
\label{sec:vmf}
CompositionalNets \cite{cvpr20} are DCNNs with an inherent robustness to partial occlusion. Their architecture resembles that of a VGG-16 network \cite{simonyan2014very}, where the fully connected head is replaced with a differentiable generative compositional model of the feature activations $p(F|y)$ and $y$ is the category of the object. 
The compositional model is defined as a mixture of von-Mises-Fisher (vMF) distributions:
\begin{align}
p(F|\Theta_y) &= \sum_m \nu_m p(F|\theta^m_y),\\
p(F|\theta^m_y) &=  \prod_{p} p(f_p|\mathcal{A}_{p,y},\Lambda), \label{eq:vmf}\\
p(f_p|\mathcal{A}_{p,y},\Lambda) &= \sum_k \alpha_{p,k,y} p(f_p|\lambda_k),\label{eq:vmf2}
\end{align}
with  $\{\nu_m \in\{0,1\},\sum_{m=1}^M \nu_m = 1 \}$. Here $M$ is the number of mixtures of compositional models and $\nu_m$ is a binary assignment variable that indicates which mixture component is active.
$\Theta_y= \{\theta^m_y = \{\mathcal{A}^m_y,\Lambda\}|m=1,\dots,M\}$ are the overall compositional model parameters and  $\mathcal{A}^m_y=\{\mathcal{A}^m_{p,y}\}$ are the parameters of the mixture components at every position $p \in \mathcal{P}$ on the 2D lattice of the feature map $F$. In particular, $\mathcal{A}^m_{p,y} = \{\alpha^m_{p,0,y},\dots,\alpha^m_{p,K,y}|\sum_{k=0}^K \alpha^m_{p,k,y} = 1\}$ are the vMF mixture coefficients, $K$ is the number of mixture components and $\Lambda = \{\lambda_k = \{\sigma_k,\mu_k \} | k=1,\dots,K \}$ are the parameters of the vMF mixture distributions:
\begin{equation}
\label{eq:vmfprob}
p(f_p|\lambda_k) = \frac{e^{\sigma_k \mu_k^T f_p}}{Z(\sigma_k)}, \norm{f_p} = 1, \norm{\mu_k}= 1, 
\end{equation}
where $Z(\sigma_k)$ is the normalization constant. The model parameters $\{\Omega,\{\Theta_y\}\}$ can be trained end-to-end as described in \cite{cvpr20}.

\textbf{Occlusion modeling.}
Following the approach presented in \cite{kortylewski2017model}, CompositionalNets can be augmented with an occlusion model. Intuitively, an occlusion model defines a robust likelihood, where at each position $p$ in the image either the object model $p(f_p|\mathcal{A}^m_{p,y},\Lambda)$ or an occluder model $p(f_p|\beta,\Lambda)$ is active:
\begin{align}
	&p(F|\Theta^m_y,\beta)\hspace{-0.075cm} =\hspace{-0.075cm} \prod_{p} p(f_p,z^m_p \hspace{0.05cm}\texttt{=}\hspace{0.05cm}0)^{1-z^m_p} p(f_p,z^m_p\hspace{0.05cm}\texttt{=}\hspace{0.05cm}1)^{z^m_p},\label{eq:occ}\\
	&p(f_p,z^m_p\hspace{0.05cm}\texttt{=}\hspace{0.05cm}1) = p(f_p|\beta,\Lambda)\hspace{0.1cm}p(z^m_p\texttt{=}1),\label{eq:occ2}\\
	&p(f_p,z^m_p\hspace{0.05cm}\texttt{=}\hspace{0.05cm}0) = p(f_p|\mathcal{A}^m_{p,y},\Lambda)\hspace{0.1cm}(1\texttt{-}p(z^m_p\texttt{=}1))\label{eq:occ3}.
\end{align}
The binary variables $\mathcal{Z}^m=\{z^m_p \in \{0,1\} | p \in\mathcal{P}\}$ indicate if the object is occluded at position $p$ for mixture component $m$. 
The occluder model is defined as a mixture model: 
\begin{align}
p(f_p|\beta,\Lambda) &= \prod_n p(f_p|\beta_n,\Lambda)^{\tau_n} \\
&=\prod_n \Big(\sum_{k} \beta_{n,k} p(f_p|\sigma_k,\mu_k)\Big)^{\tau_n},
\end{align}
where \{$\tau_n \in \{0,1\},\sum_n \tau_n =1\}$ indicates which component of the occluder model best explains the data.
The parameters of the occluder model $\beta_n$ can be learned in an unsupervised manner from clustered features of random natural images that do not contain any object of interest. 
\begin{figure}
\centering
\includegraphics[height=3.6cm]{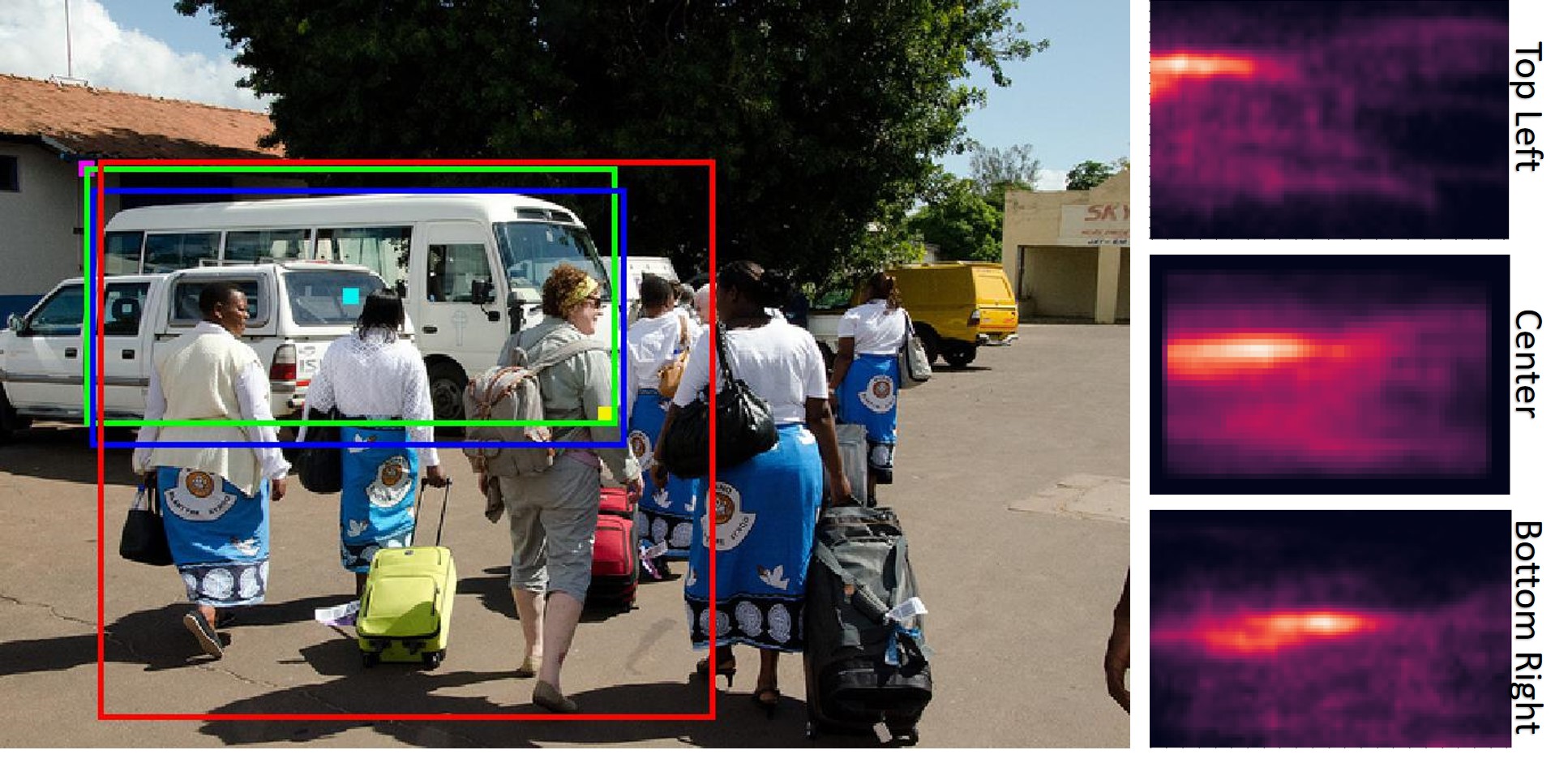}
\caption{Example of robust bounding box voting results. Blue box: ground truth; red box: bounding box by Faster R-CNN; green box: bounding box generated by robustly combining voting results.
Our proposed part-based voting mechanism generates probability maps (right) for the object center (cyan point), the top left corner (purple point) and the bottom right corner (yellow point) of the bounding box. }
\label{fig:corner}
\end{figure}

\subsection{Detection with Robust Bounding Box Voting}
\label{sec:detect}

A natural way of generalizing CompositionalNets to object detection is to combine them with RPNs. However, our experiments in Section \ref{sec:base} show that RPNs cannot reliably localize strongly occluded objects. 
Figure \ref{fig:rpn} illustrates this limitation by depicting the detection results of Faster R-CNN trained with CutOut \cite{cutout} (red box) and a combination of RPN+CompositionalNet (yellow box). 
We propose to address this limitation by introducing a robust part-based voting mechanism to predict the bounding box of an object based on the visible object parts (green box).

\begin{figure}
    \centering
    \includegraphics[width=8.7cm]{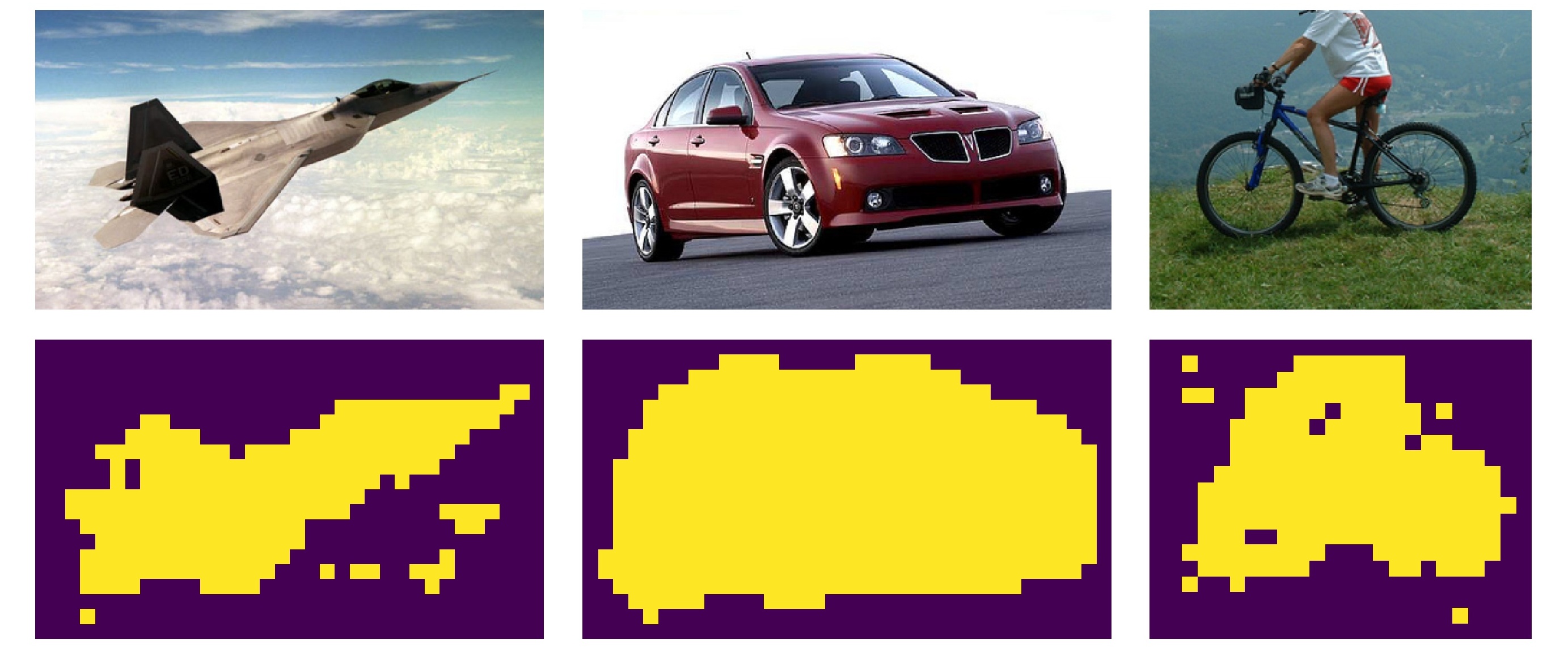}
    \caption{Context segmentation results.
    A standard CompositionalNet learns a joint representation of the image including the context. Our context-aware CompositionalNet will disentangle the representation of the context from that of the object based on the illustrated segmentation masks.}
    \label{fig:seg}
\end{figure}

\textbf{CompositionalNets with detection layer.}
CompositionalNets as introduced in \cite{cvpr20} are part-based object representations. 
In particular, the object model $p(F|\Theta_y)$ is decomposed into a mixture of compositional models $p(F|\theta^m_y)$, where each mixture component represents the object class $y$ from a different pose \cite{cvpr20}. 
During inference, each mixture component accumulates votes from part models $p(f_p|A_{p,y})$ across different spatial positions $p$ of the feature map $F$.
Note that CompositionalNets are learned from images that are cropped based on the bounding box of the object \cite{cvpr20}. By making the object centered in the image (see Figure \ref{fig:seg}), each mixture component $p(F|\theta^m_y)$ can be thought of as accumulating votes from the part models for the object being in the center of the feature map. 

Based on this intuition, we generalize CompositionalNets to object detection by introducing a detection layer that accumulates votes for the object center over all positions $p$ in the feature map $F$.
In order to achieve this, we propose to compute the object likelihood by scanning.
Thus, we shift the feature map w.r.t. the object model along all points $p$ from the 2D lattice of the feature map.
This process will generate a spatial likelihood map:
\begin{equation}
    R = \{p(F_{p}|\Theta_y)| p \in \mathcal{P}\}, 
\end{equation}
where $F_{p}$ denotes the feature map centered at the position $p$. 
Using this generalization we can perform object localization by selecting all maxima in $R$ above a threshold $t$ after non-maximum suppression. 
Our proposed detection layer can be implemented efficiently with modern hardware using convolution-like operations. 

\begin{figure}
\centering
\includegraphics[height=3.6cm]{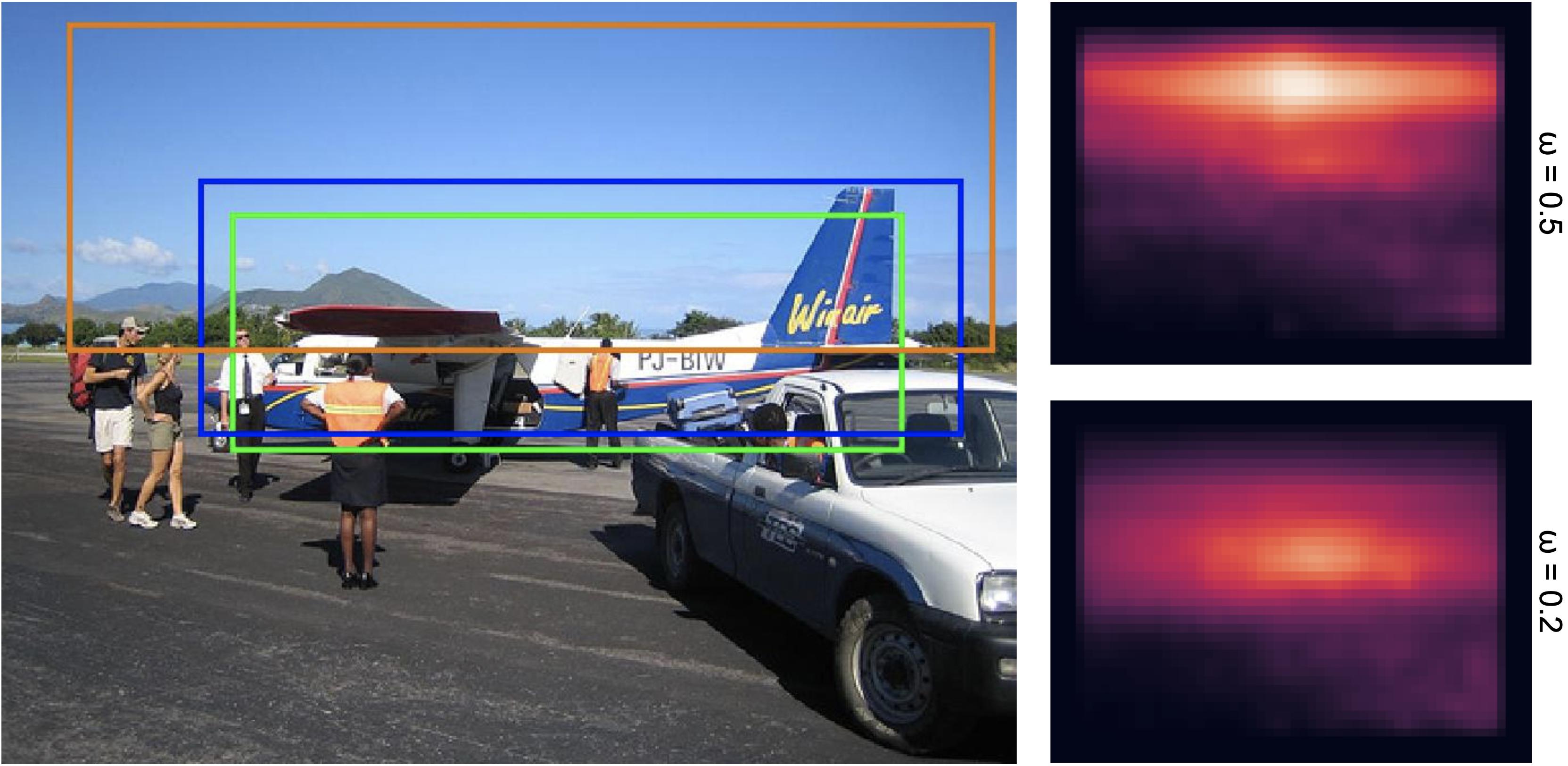}
\caption{Influence of context in aeroplane detection under occlusion. Blue box: ground truth; orange box: bounding box by CompositionalNets ($\omega=0.5$); green box: bounding box by Context-Aware CompositionalNets ($\omega=0.2$).
Probability maps of the object center are on the right. 
Note how reducing the influence of the context improves the localization response.
}
\label{fig:context_bad}
\end{figure}
\textbf{Robust bounding box voting.} 
While CompositionalNets can be generalized to localize partially occluded objects using our proposed detection layer, estimating the bounding box of an object under occlusion is more difficult because a significant amount of the object might not be visible (Figure \ref{fig:corner}). 
We propose to solve this problem by generalizing the part-based voting mechanism in CompositionalNets to vote for the bounding box corners in addition to the object center.
In particular, we learn additional mixture components that model the expected feature activations $F$ around bounding box corners $p(F_{p}|\Theta_y^c)$, where $c=\{ct,bl,tr\}$ are the object center $ct$ and two opposite bounding box corners $\{bl,tr\}$.
Figure \ref{fig:corner} illustrates the spatial likelihood maps $R^c$ of all three models.
We generate a bounding box using the two points that have maximal likelihood.
Note how the bounding boxes can be localized accurately despite large parts of the object being occluded.
We discuss how the parameters of all models can be learned jointly in an end-to-end manner in Section \ref{sec:e2e}.

\subsection{Context-aware CompositionalNets}
\label{sec:context}
CompositionalNets, as well as standard DCNNs, do not separate the representation of the context from the object.
The context can be useful for recognizing objects due to biases, e.g. aeroplanes are often surrounded by blue sky. Relying too strongly on context can be misleading when objects are strongly occluded (Figure \ref{fig:context_bad}),
since the detection thresholds must be lowered under strong occlusion. This in turn increases the influence of the objects' context and leads to false-positive detection in regions with no object. 
Hence, it is important to have control over the influence of contextual cues on the detection result.

In order to gain control over the influence of context, we propose a Context-aware CompositionalNets (CA-CompositionalNets), which separates the representation of the context from the object in the original CompositionalNets by representing the feature map $F$ as a mixture of two models:
\begin{align}
    	p(f_p|\mathcal{A}^m_{p,y},\chi^m_{p,y},\Lambda)=
    	&\omega \hspace{0.1cm}p(f_p|\chi^m_{p,y},\Lambda) + \\
    	&(1-\omega) p(f_p|\mathcal{A}^m_{p,y},\Lambda).
\end{align}{}
Here,  $\chi^m_{p,y}$ are the parameters of the context model that is defined to be a mixture of vMF likelihoods (Equation \ref{eq:vmf2}).
The parameter $\omega$ is a prior that controls the trade-off between context and object, which is fixed a priori at test time. 
Note that setting $\omega=0.5$ retains the original CompositionalNet as proposed in \cite{cvpr20}.
Figure \ref{fig:context_bad} illustrates the benefits of reducing the influence of the context on the detection result under partial occlusion.
The context parameters $\chi^m_{p,y}$ and object parameters $\mathcal{A}^m_{p,y}$ can be learned from the training data using maximum likelihood estimation. 
However, this presumes an assignment of the feature vectors $f_p$ in the training data to either the context or the object.

\textbf{Context segmentation.} Therefore, we propose to segment the training images into context and object based on the available bounding box annotation. Here, our assumption is that any feature that has a receptive field outside of the scope of the bounding boxes would be considered as a part of the context. 
We first randomly extract features that are considered to be context into a population during training. 
Then, we cluster the population using K-means++ algorithm\cite{k-means++} and receive a dictionary of context feature centers $E=\{e_q \in \mathbb{R}^D |q= 1,\dots,Q \}$. 
We apply a threshold on the cosine similarity $s(E,f_p)=\max_q [(e_q^T f_p)/(\norm{e_q}\norm{f_p})]$ to segment the context and the object in any given training image (Figure \ref{fig:seg}).

\subsection{Training Context-Aware CompositionalNets}
\label{sec:e2e}
We train our proposed CA-CompositionalNet including the robust bounding box voting mechanism jointly end-to-end using backpropagation. 
Overall, the trainable parameters of our models are $T^c=\{\Omega,\Lambda,\{\Theta^c_y\},\{\chi^c_y\}\}$ where $c\in\{ct,bl,tr\}$.
The loss function has three main objectives: 
optimizing the parameters of the generative compositional model such that it can explain the data with maximal likelihood ($\mathcal{L}_{g}$), while also localizing ($\mathcal{L}_{detect}$) and classifying ($\mathcal{L}_{cls}$) the object accurately in the training images.
While $\mathcal{L}_g$ is learned from images $\hat{I^c}$ with feature maps $F^c$ that are centered at $c\in\{c,bl,tr\}$, the other losses are learned from unaligned training images $I$ with feature maps $F$.

\textbf{Training Classification with Regularization. } We optimize the parameters jointly using SGD: 
\begin{align}
    \mathcal{L}_{cls}(y,y') =  & \mathcal{L}_{class}(y,y') + \mathcal{L}_{weight}(\Omega) 
\end{align}
where $\mathcal{L}_{class}(y,y')$ is the cross-entropy loss between the network output $y'=\psi(I,\Omega)$ and the true class label $y$. We use a temperature $\mathcal{T}$ in the softmax classifier: $f(y)_i=\frac{e^{y_i\cdot \mathcal{T}}}{\Sigma_{i}e^{y_i\cdot \mathcal{T}}}$. 
$\mathcal{L}_{weight} = \norm{\Omega}^2_2$ is a weight regularization on the DCNN parameters.

\textbf{Training the generative context-aware CompositionalNet.} 
The overall loss function for training the parameters of the generative context-aware model is composed of two terms:
\begin{align}
    \mathcal{L}_{g}(F^c,T) =  
     \mathcal{L}_{vmf}(F^c,\Lambda) \\
     + \sum_c \sum_p \mathcal{L}_{con}(f^c_p,\mathcal{A}^c_y,\chi^c_y)
\end{align}
In order to avoid the computation of the normalization constants $\{Z[\sigma_k]\}$, we assume that the vMF variances
$\{\sigma_k\}$ are constant. Under this assumption, the vMF parameters $\{\mu_k\}$ can be optimized with the loss
$\mathcal{L}_{vmf}(F,\Lambda) = \mathcal{C}\sum_p \min_k \mu_k^T f_p$, where $\mathcal{C}$ is a constant factor \cite{cvpr20}. The parameters of the context-aware model $\mathcal{A}^c_y$ and $\chi^c_y$ are learned by optimizing the context loss:
\begin{align}
    \mathcal{L}_{con}(f_p,\mathcal{A}^c_y,\chi^c_y) = &\pi_p \mathcal{L}_{mix}(f_p,\mathcal{A}^c_{p,y})
\end{align}
where $\pi_p \in \{0,1\}$ is a context assignment variable that indicates if a feature vector $f_p$ belongs to the context or to the object model. We estimate the context assignments a priori using segmentation as described in Section \ref{sec:context}. Given the assignments we can optimize the model parameters $\mathcal{A}^c_{p,y}$ by minimizing \cite{kortylewski2019compositional}:
\begin{equation}
    \mathcal{L}_{mix}(F,\mathcal{A}^c_y) =\hspace{-.05cm}\texttt{-}\hspace{-.1cm}\sum_p\hspace{-0.05cm}(1\texttt{-}z^\uparrow_{p}) \log \hspace{-0.05cm}\Big[\hspace{-0.05cm}\sum_k\hspace{-0.1cm}\alpha^{m^\uparrow,c}_{p,k,y}p(f_p|\lambda_k)\Big]
\end{equation}
The context parameters $\chi^c_{p,y}$ can be learned accordingly. Here, $z^\uparrow_{p}$ and $m^\uparrow$ denote the variables that were inferred in the forward process. Note that the parameters of the occluder model are learned a priori and then fixed. 

\textbf{Training for localization and bounding box localization.} 
We denote the normalized response map of the ground truth class as 
$X^{c}\in\mathbb{R}^{H\times W}$
and the ground truth annotation as $\bar{X}^c\in\mathbb{R}^{H\times W}$. 
The elements of the response map are computed as:
\begin{equation}
x_{p}^c = \frac{x_{p, \hat{m}}}{\sum_{p}x_{p, \hat{m}}}, \hat{m}=\argmax_m \max_{p}
p(f_{p}|\mathcal{A}^m_{p,y},\chi^m_{p,y},\Lambda).
\end{equation}
The ground truth map $\bar{X}^c$ is a binary map where the ground truth position is set to $X^c(c)=1$ and all other entries are set to zero. 
The detection loss is then defined as: 
\begin{equation}
\mathcal{L}_{detect}(X^c,\bar{X}^c,F,T^c) = 1 -\frac{2\cdot \Sigma_{p} (x^c_{p} \cdot \bar{x}^c_p)}{\sum_{p} x^c_p + \sum_{p}  \bar{x}^c_{p}}
\end{equation}

\textbf{End-to-end training.} We train all parameters of our model end-to-end with backpropagation. The overall loss function is:
\begin{align}
    \label{eq:loss}
    \mathcal{L} = \mathcal{L}_{cls}(y, y') +
    \sum_c \big( \epsilon_1 \mathcal{L}_g(F^c,T^c) \\
    + \epsilon_2 \mathcal{L}_{detect}(X^c,\bar{X}^c,F,T^c) \big) 
\end{align}
$\epsilon_1, \epsilon_2$ control  the  trade-off  between  the  loss terms.
The optimization process is discussed in more detail in Section \ref{sec:exp}.

\begin{table*}[t]
\tabcolsep=0.11cm
\begin{tabular}{cV{2.5}c V{2.5} ccc V{2.5} ccc V{2.5} ccc V{2.5} c}
                      & FG L0 & \multicolumn{3}{c}{FG L1} & \multicolumn{3}{V{2.5}c}{FG L2} & \multicolumn{3}{V{2.5}cV{2.5}}{FG L3} & \multicolumn{1}{c}{Mean} \\
                      \cline{2-11}
method                & BG L0 & BG L1   & BG L2  & BG L3  & BG L1   & BG L2  & BG L3  & BG L1   & BG L2  & BG L3 & -- \\
\hline
Faster R-CNN           &{\bf 98.0}     &88.8       &85.8       &83.6       &72.9       &66.0        &60.7       &46.3      &36.1       &27.0       &66.5        \\
Faster R-CNN with reg. &97.4     &{\bf 89.5}       &86.3       &89.2       &76.7       &70.6        &67.8       &54.2      &45.0       &37.5       &71.1        \\
\hline
CA-CompNet via RPN $\omega$=$0.5$ & 74.2     & 68.2    & 67.6    & 67.2   & 61.4     & 60.3     & 59.6   & 46.2    & 48.0    & 46.9   & 60.0 \\
CA-CompNet via RPN $\omega$=$0$ & 73.1    & 67.0    & 66.3    & 66.1   & 59.4    & 60.6    & 58.6   & 47.9    & 49.9    & 46.5   & 59.6 \\
\hline

CA-CompNet via BBV $\omega$=$0.5$ & 91.7	& 85.8	& 86.5	& 86.5	& 78.0	& 77.2	& 77.9	& 61.8	& 61.2	& 59.8	& 76.6 \\
CA-CompNet via BBV $\omega$=$0.2$ & 92.6  & 87.9    & 88.5    & {\bf 88.6}   & 82.2    & {\bf 82.2}    & {\bf 81.1}   & 71.5    & {\bf 69.9}    & {\bf 68.2}   & 81.3 \\
CA-CompNet via BBV $\omega$=$0$ & 94.0    & 89.2    & {\bf 89.0}    & 88.4   & {\bf 82.5}    & 81.6    & 80.7   & {\bf 72.0}    & 69.8    & 66.8   & {\bf 81.4} 
\end{tabular}
\caption{Detection results on the OccludedVehiclesDetection dataset under different levels of occlusions (BBV as in Bounding Box Voting). All models trained on PASCAL3D+ unoccluded dataset except Faster R-CNN with reg. was trained with CutOut. The results are measured by correct AP(\%) @IoU0.5, which means only corrected classified images with $IoU>0.5$ of first predicted bounding box are treated as true-positive. Notice with $\omega=0.5$, context-aware model reduces to a CompositionalNet as proposed in \cite{cvpr20}.}
\label{tab:p3d}
\end{table*}

\begin{table}[]
\tabcolsep=0.11cm
\begin{tabular}{cV{2.5}ccV{2.5}ccc}
 & \multicolumn{2}{c}{light occ.} & \multicolumn{3}{V{2.5}c}{heavy occ.} \\
       \cline{2-6}
method & L0 & L1 & L2 & L3 & L4\\
       \hline
Faster R-CNN                        & 81.7	&66.1	&59.0 	&40.8	&24.6 \\
Faster R-CNN with reg.              & 84.3	&71.8	&63.3	&45.0 	&33.3 \\
Faster R-CNN with occ.              & 85.1	&76.1	&66.0	&50.7	&45.6 \\
       \hline
CA-CompNet via RPN $\omega$=$0$       & 62.0 & 55.0 & 49.7 & 45.4 & 38.6 \\
       \hline
       
CA-CompNet via BBV $\omega$=$0.5$     & 83.5 & 77.1 & 70.8 & 51.7 & 40.4 \\
CA-CompNet via BBV $\omega$=$0.2$      & 88.7 & 82.2 & {\bf 77.8} & {\bf 65.4} & {\bf 59.6} \\
CA-CompNet via BBV $\omega$=$0$       & {\bf 91.8} & {\bf 83.6} & 76.2 & 61.1 & 54.4
\end{tabular}
\caption{Detection results on OccludedCOCO Dataset, measured by AP(\%) @IoU0.5. All models are trained on PASCAL3D+ dataset, Faster R-CNN with reg. is trained with CutOut and Faster R-CNN with occ. is trained with images in same dataset but occluded by all levels of occlusion with the same set of occluders.}
\label{tab:coco}
\end{table}

\begin{figure}
    \centering
    \includegraphics[height=5cm]{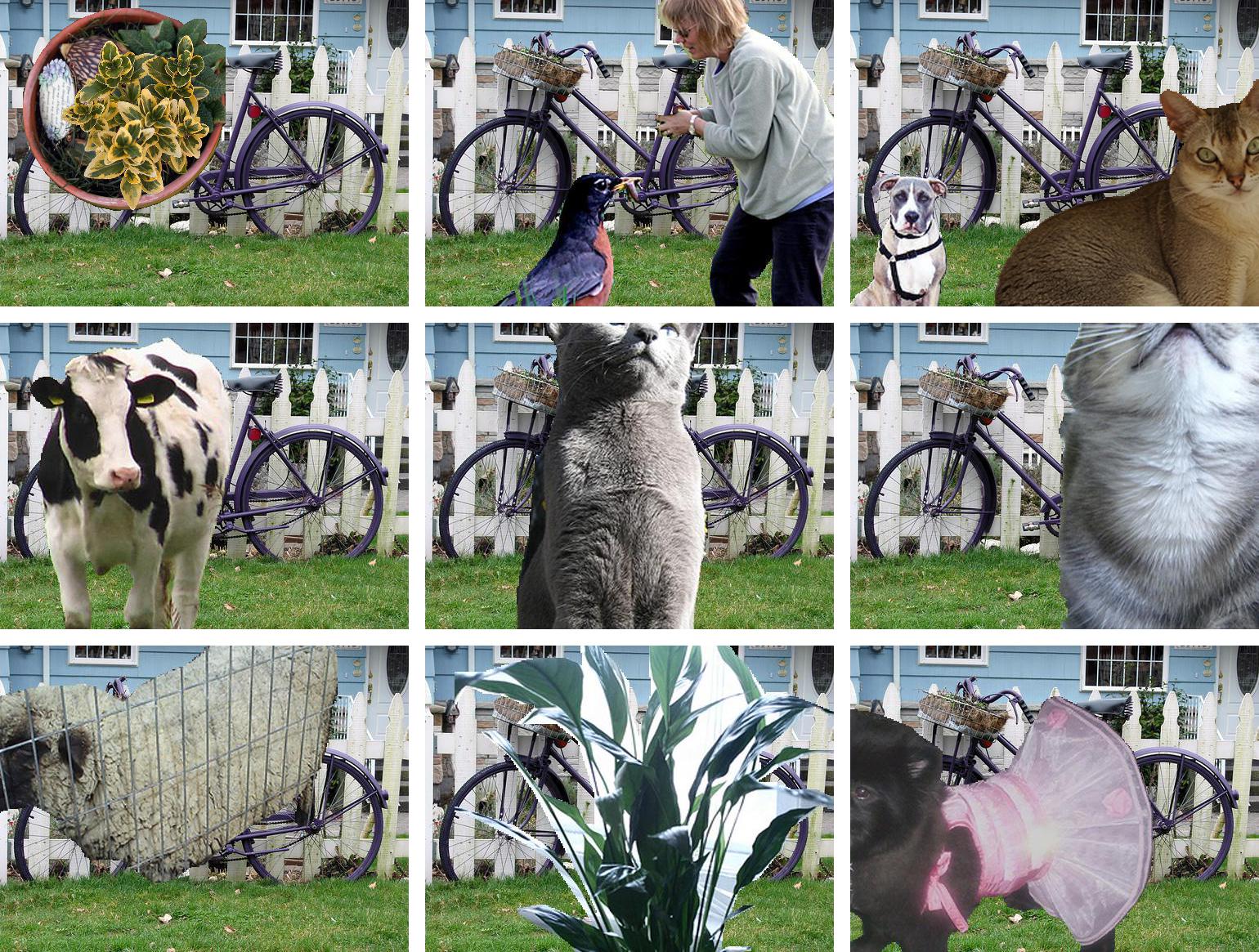}
    \caption{Example of images in OccludedVehiclesDetection dataset. Each row shows increasing amounts of context occlusion, whereas each column shows increasing amounts of object occlusion.}
    \label{fig:dataset}
\end{figure}

\section{Experiments}
\label{sec:exp}
We perform experiments on object detection under artificially-generated and real-world occlusion.

\textbf{Datasets.} 
While it is important to evaluate algorithms on real images of partially occluded objects, simulating occlusion enables us to quantify the effects of partial occlusion more accurately.
Inspired by the success of datasets with artificially-generated occlusion in image classification \cite{occdata}, we propose to generate an analogous dataset for object detection.
In particular, we build on the PASCAL3D+ dataset, which contains 12 classes of unoccluded objects. 
We synthesize an \textit{OccludedVehiclesDetection} dataset similar to the dataset proposed in \cite{occdata} for classification, which contains 6 classes of vehicles at a fixed scale (224 pixels) and various levels of occlusion. 
The occluders, which include humans, animals and plants, are cropped from the MS-COCO dataset \cite{lin2014microsoft}.
In an effort to accurately depict real-world occlusions, we superimpose the occluders onto the object, such that the occluders are placed not only inside the bounding box of the objects, but also on the background. We generate the dataset in a total of 9 occlusion levels along two dimensions. We define three levels of object occlusion: FG-L1: 20-40\%, FG-L2: 40-60\% and FG-L3: 60-80\% of the object area occluded. Furthermore, we define three levels of context occlusion around the object: BG-L1: 0-20\%, BG-L2: 20-40\% and BG-L3: 40-60\% of the context area occluded. An example of occlusion levels are showed in Figure \ref{fig:dataset}.

In order to evaluate the tested models on real-world occlusions, we test them on a subset of the MS-COCO dataset.
In particular, we extract the same classes of objects and scale as in the OccludedVehiclesDetection dataset from the MS-COCO dataset. We select occluded images and manually separate them into two groups: light occlusion (2 sub-levels) and heavy occlusions (3 sub-levels), with increasing occlusion levels. 
This dataset is built from images in both Training2017 and Val2017 set of MS-COCO due to a limited amount of heavily occluded objects in MS-COCO Dataset. The light occlusion set contains 2890 images, and the heavy occlusion set contains 788 images. We term this dataset \textit{OccludedCOCO}.

{\bf Evaluation.} In order to exclusively observe the effects of foreground and background occlusions on various models, we only consider the occluded object in the image for evaluation. Evidently, for the majority of the dataset, there is often only one object of a particular class that is present in the image. This enables us to quantify the effects of levels of occlusions in the foreground and background on the accuracy of the model predictions. Thus, the means of object detection evaluation must be altered for our proposed occlusion dataset. Given any model, we only evaluate the bounding box proposal with the highest confidence given by the classifier via IoU at 50\%. 

{\bf Runtime. } The convolution-like detection layer has an inference time of 0.3s per image.

\textbf{Training setup.}
We implement the end-to-end training of our CA-CompositionalNet with the following parameter settings: training minimizes loss described in Equation \ref{eq:loss}, with $\epsilon_1=0.2$ and $\epsilon_2=0.4$. We applied the Adam Optimizer \cite{kingma2014adam} with various learning rates of $lr_{vgg}=2\cdot 10^{-6}$, $lr_{vc}=2 \cdot 10^{-5}$, $lr_{mixture \ model}=5 \cdot 10^{-5}$ and $lr_{corner \ model}=5 \cdot 10^{-5}$ on different parts of CompositionalNets. The model is trained for a total of 2 epochs with 10600 iteration per epoch. The training costs in total of 3 hours on a machine with 4 NVIDIA TITAN Xp GPUs.

Faster R-CNN is trained for 30000 iterations, with a learning rate, $lr=1\cdot 10^{-3}$, and a learning rate decay, $lr_{decay}=0.1$. Specifically, the pretrained VGG-16 \cite{simonyan2014very} on the ImageNet dataset \cite{deng2009imagenet} was modified in its fully-connected layer to accommodate the experimental settings. In the experiment on OccludedCOCO, we set the threshold of Faster R-CNN to 0, preventing the occluded targets to be ignored due to low confidence and guarantees at least one proposal in the required class. 

\subsection{Object Detection under Simulated Occlusion}
\label{sec:base}

\begin{figure}
\begin{subfigure}{.33\textwidth}
  \centering
  \includegraphics[height=1.8cm]{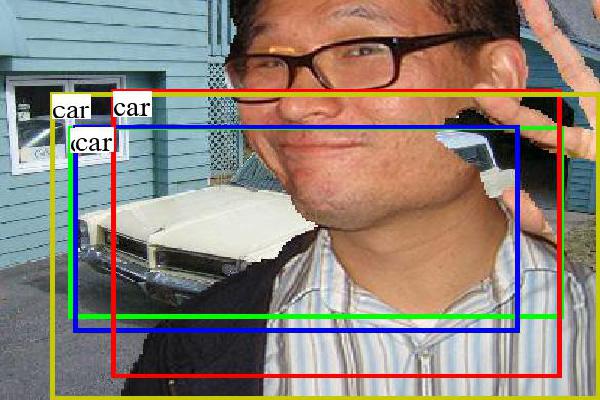}
  \label{fig:sfig1}
\end{subfigure}%
\begin{subfigure}{.33\textwidth}
  \centering
  \includegraphics[height=1.8cm]{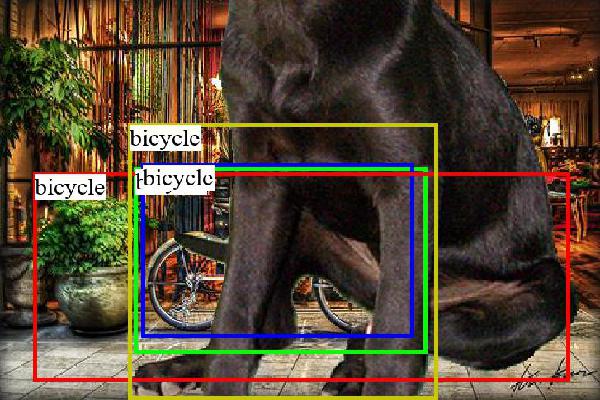}
  \label{fig:sfig2}
\end{subfigure}
\begin{subfigure}{.31\textwidth}
  \centering
  \includegraphics[height=1.8cm]{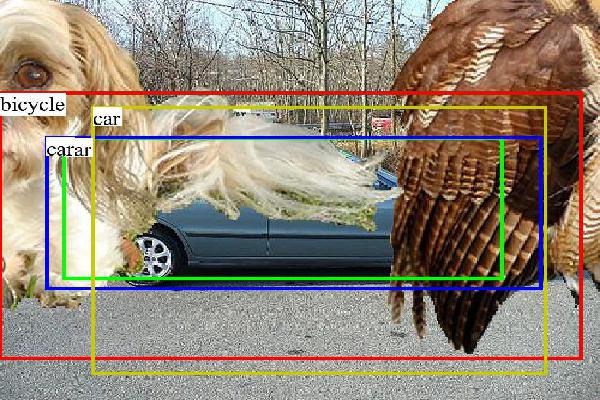}
  \label{fig:sfig3}
\end{subfigure}
\begin{subfigure}{.33\textwidth}
  \centering
  \includegraphics[height=1.8cm]{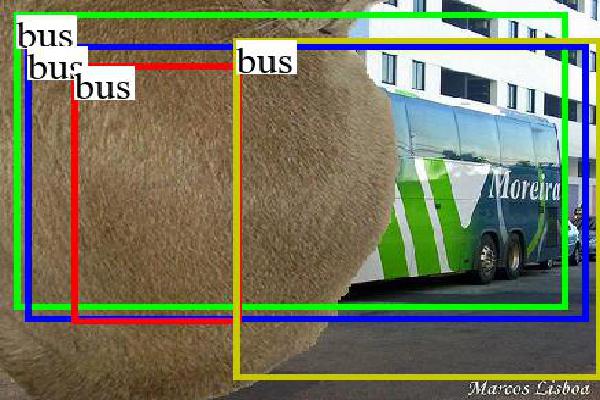}
  \label{fig:sfig4}
\end{subfigure}%
\begin{subfigure}{.33\textwidth}
  \centering
  \includegraphics[height=1.8cm]{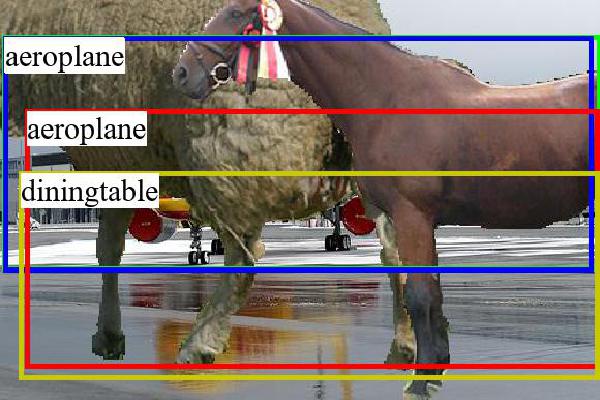}
  \label{fig:sfig5}
\end{subfigure}
\begin{subfigure}{.31\textwidth}
  \centering
  \includegraphics[height=1.8cm]{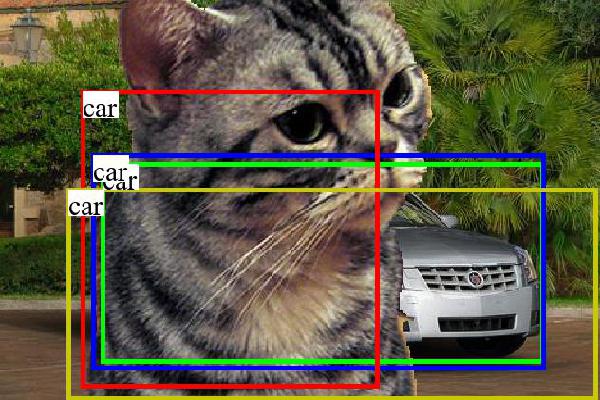}
  \label{fig:sfig6}
\end{subfigure}
\caption{Selected examples of detection results on the  OccludedVehiclesDetection dataset. All of these 6 images are the heaviest occluded images (foreground level 3, background level 3). Blue box: ground truth; green box: proposals of CA-CompositionalNet via BB Voting; yellow box: proposals of CA-CompositionalNet via RPN; red box: proposals of Faster R-CNN.}
\label{fig:VisualPASCAL}
\end{figure}

Table \ref{tab:p3d} shows the results of the tested models on the OccludedVehiclesDetection dataset (see Figure \ref{fig:VisualPASCAL} for qualitative results). 
The models are trained on the images from the original PASCAL3D+ dataset with unoccluded objects. 

\textbf{Faster R-CNN.} 
As we evaluate the performance of the Faster R-CNN, we observe that under low levels of occlusion, the neural network performs well. In mid to high levels of occlusions, however, the neural network fails to detect the objects robustly. 
When trained with strong data augmentation in terms of partial occlusion using CutOut \cite{cutout}, the detection performance increases under strong occlusion. However, the model still suffers from a $59.9\%$ drop in performance on strong occlusion, compared to the non-occlusion setup.
We suspect that the inaccurate prediction is due to two major factors: 1) The Region Proposal Network (RPN) in the Faster R-CNN is not able to predict accurate proposals of objects that are heavily occluded. 2) The VGG-16 classifier cannot successfully classify valid object regions under heavy occlusion.

We proceed to investigate the performance of the region proposals on occluded images. 
In particular, we replace the VGG-16 classifier in the Faster R-CNN with a standard CompositionalNet classifier \cite{cvpr20}, which is expected to be more robust to occlusion. 
From the results in Table 1, we observe two phenomena: 1) In high levels of occlusion, the performance is better than Faster R-CNN. Thus, the CompositionalNet generalizes to heavy occlusions better than the VGG-16 classifier. 2) In low levels of occlusion, the performance is worse than Faster R-CNN. 
The proposals generated by the RPN seem to be not accurate enough to be correctly classified, as CompositionalNets are high-precision models and require a precise alignment of the bounding box to the object center.

\textbf{Effect of robust bounding box voting.} Our approach of estimating corners of the bounding box substantially improves the performance of the CompositionalNets, in comparison to the RPN. This further validates our conclusion that the CompositionalNet classifier requires precise proposals to classify objects correctly with partial occlusions.

\textbf{Effect of context-aware representation.} With $\omega=0.5$, we observe that the precision of the detection decreases. Furthermore, the performance between $\omega=0.5$ and $\omega=0$ follows a similar trend over all three levels of foreground occlusions: the performance decreases as the level of background occlusion increases from BG-L1 to BG-L3. This further confirms our understanding of the effects of the context as a valuable source of information in object detection.

\begin{figure}
\begin{subfigure}{.40\textwidth}
  \centering
  \includegraphics[height=3.25cm]{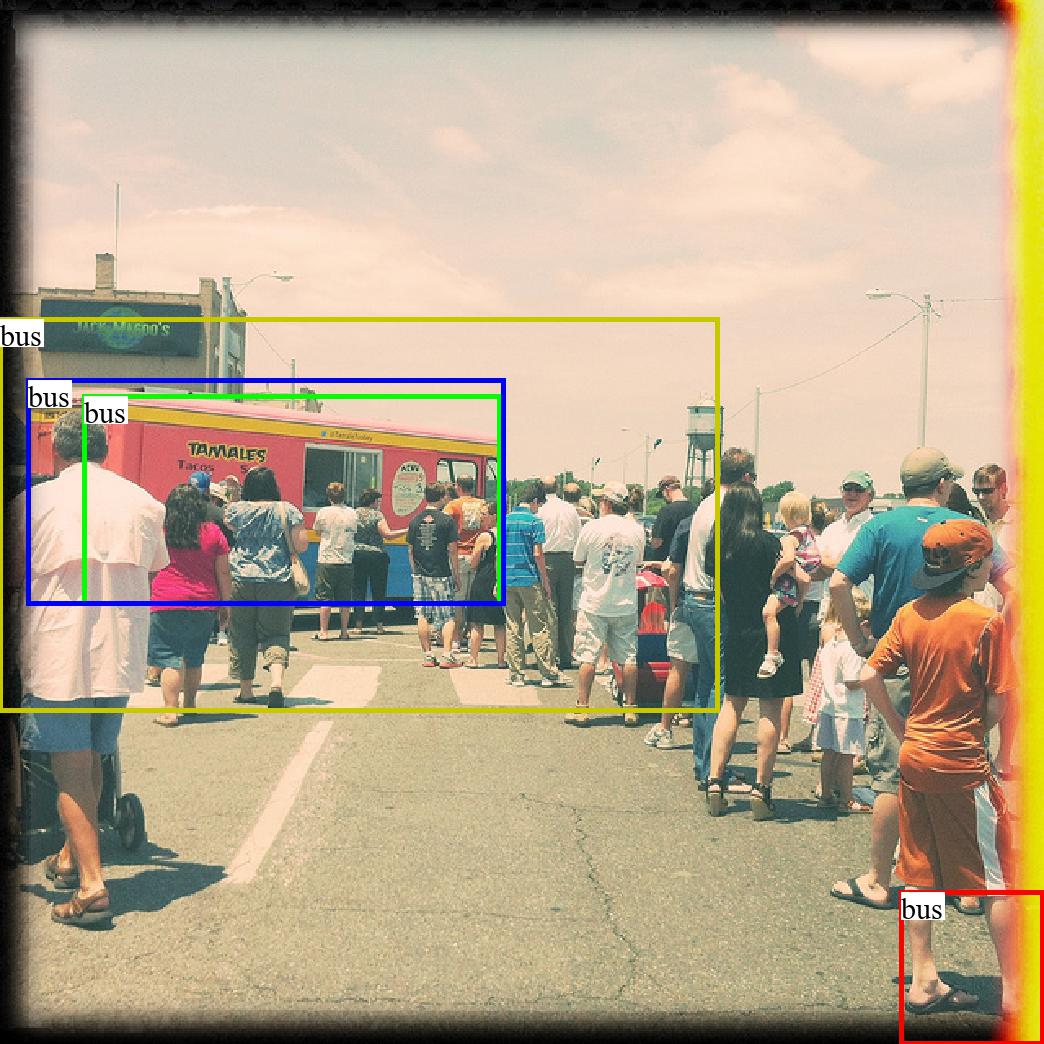}
  \label{fig:sfig1}
\end{subfigure}%
\begin{subfigure}{.28\textwidth}
  \centering
  \includegraphics[height=3.3cm]{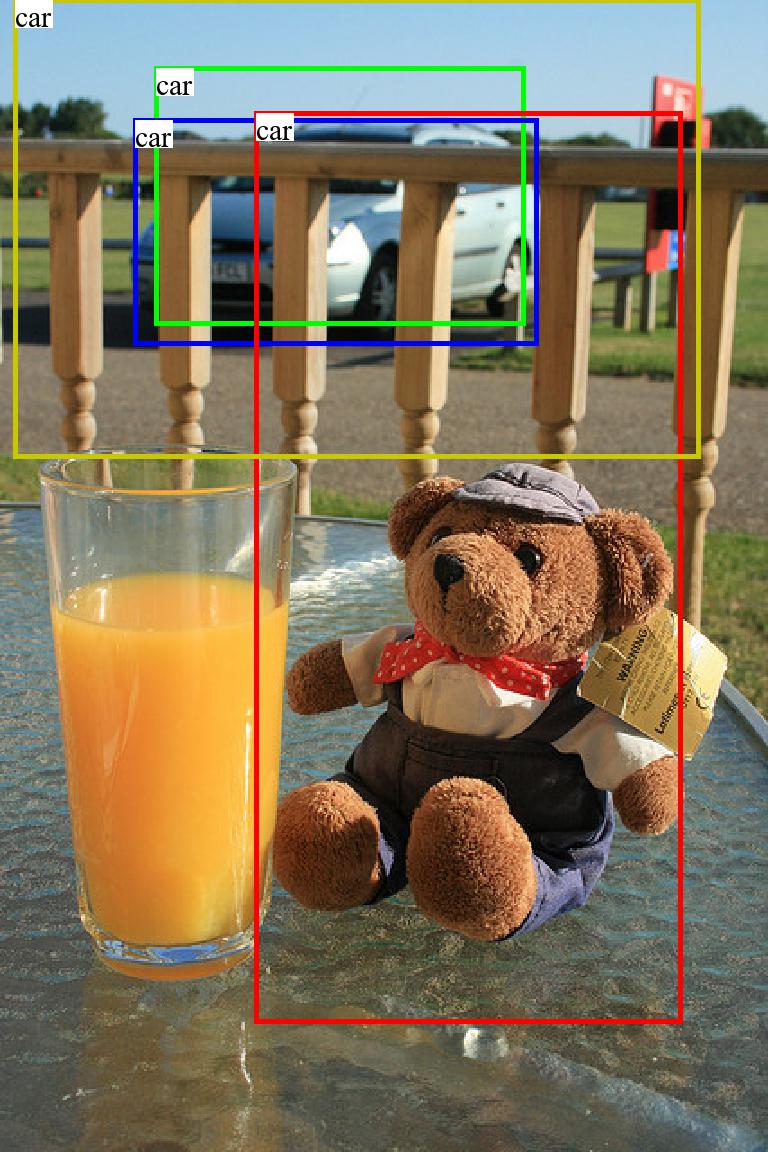}
  \label{fig:sfig2}
\end{subfigure}
\begin{subfigure}{.28\textwidth}
  \centering
  \includegraphics[height=3.3cm]{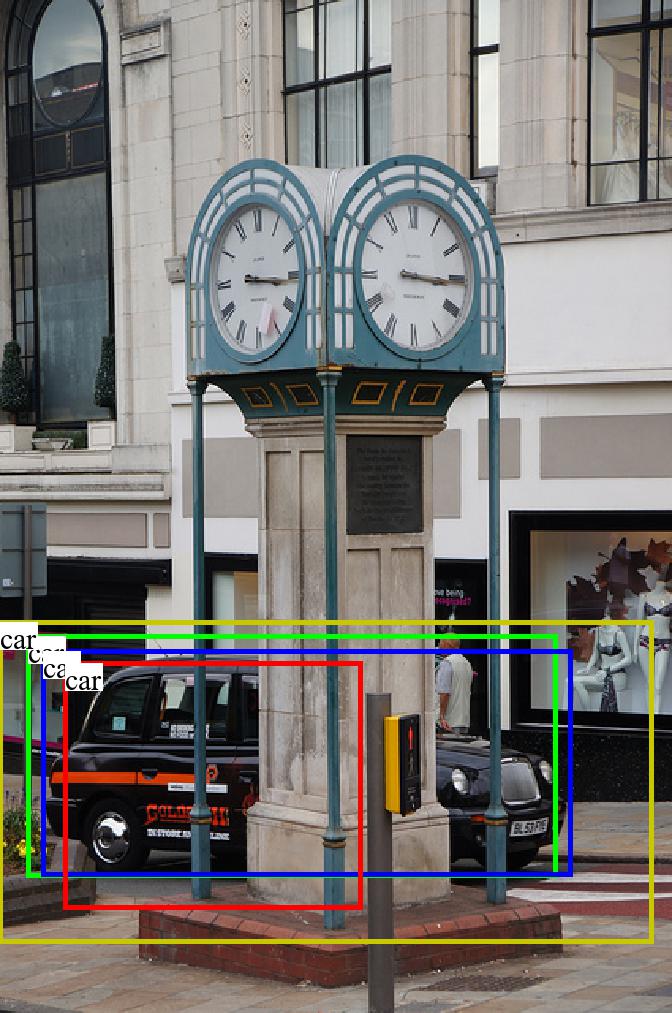}
  \label{fig:sfig3}
\end{subfigure}
\begin{subfigure}{.40\textwidth}
  \centering
  \includegraphics[height=1.75cm]{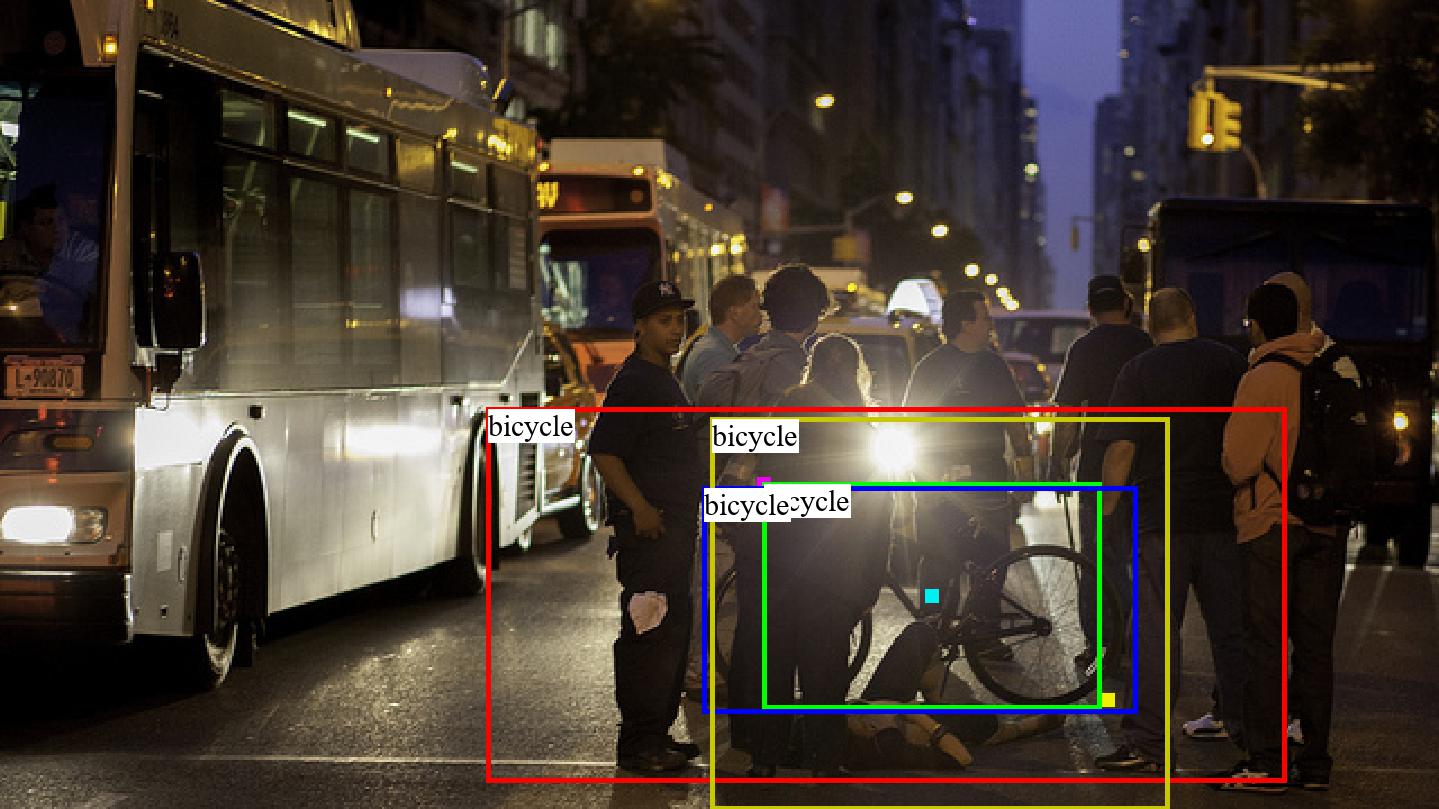}
  \label{fig:sfig4}
\end{subfigure}%
\begin{subfigure}{.29\textwidth}
  \centering
  \includegraphics[height=1.75cm]{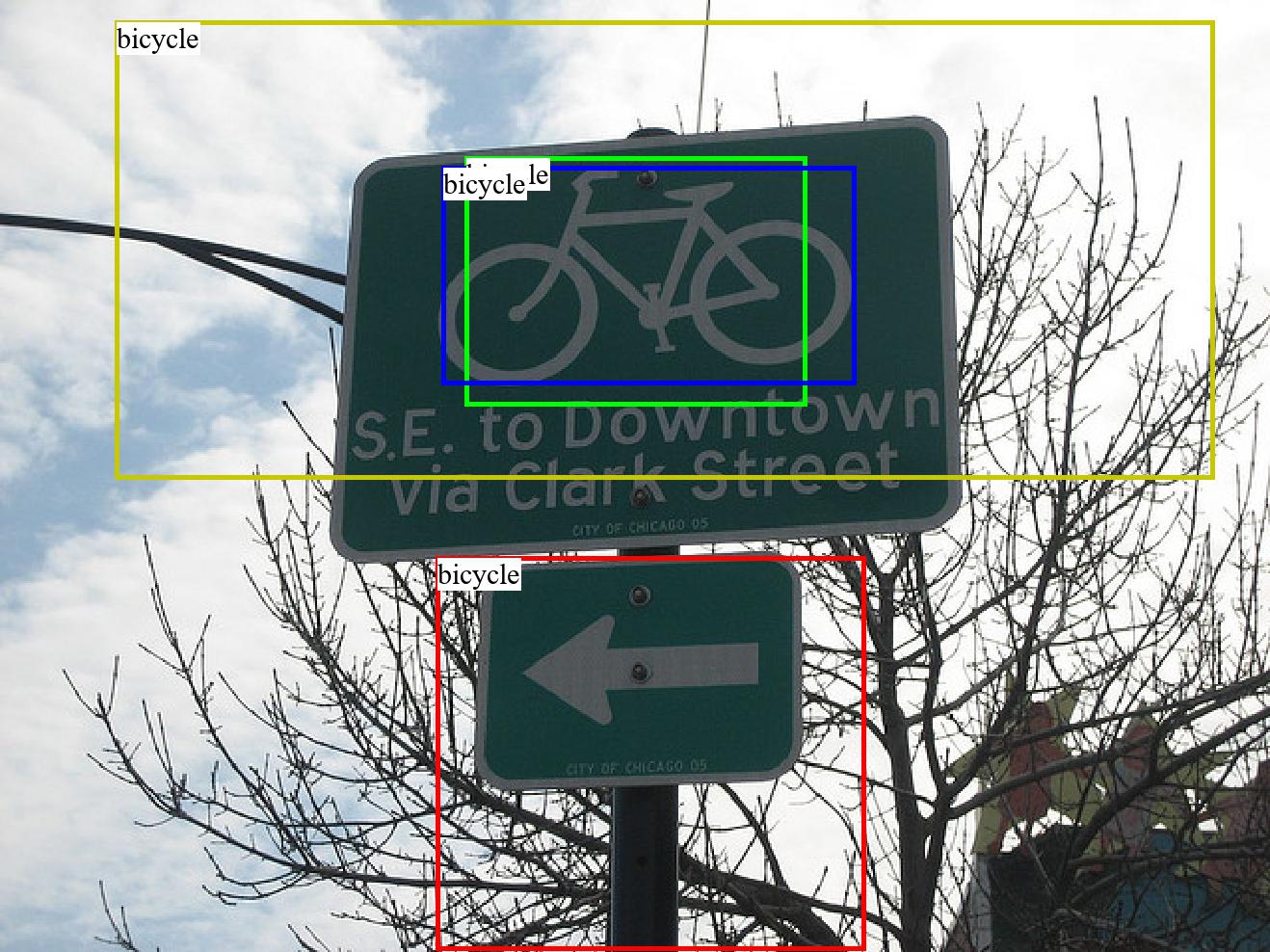}
  \label{fig:sfig5}
\end{subfigure}
\begin{subfigure}{.29\textwidth}
  \centering
  \includegraphics[height=1.75cm]{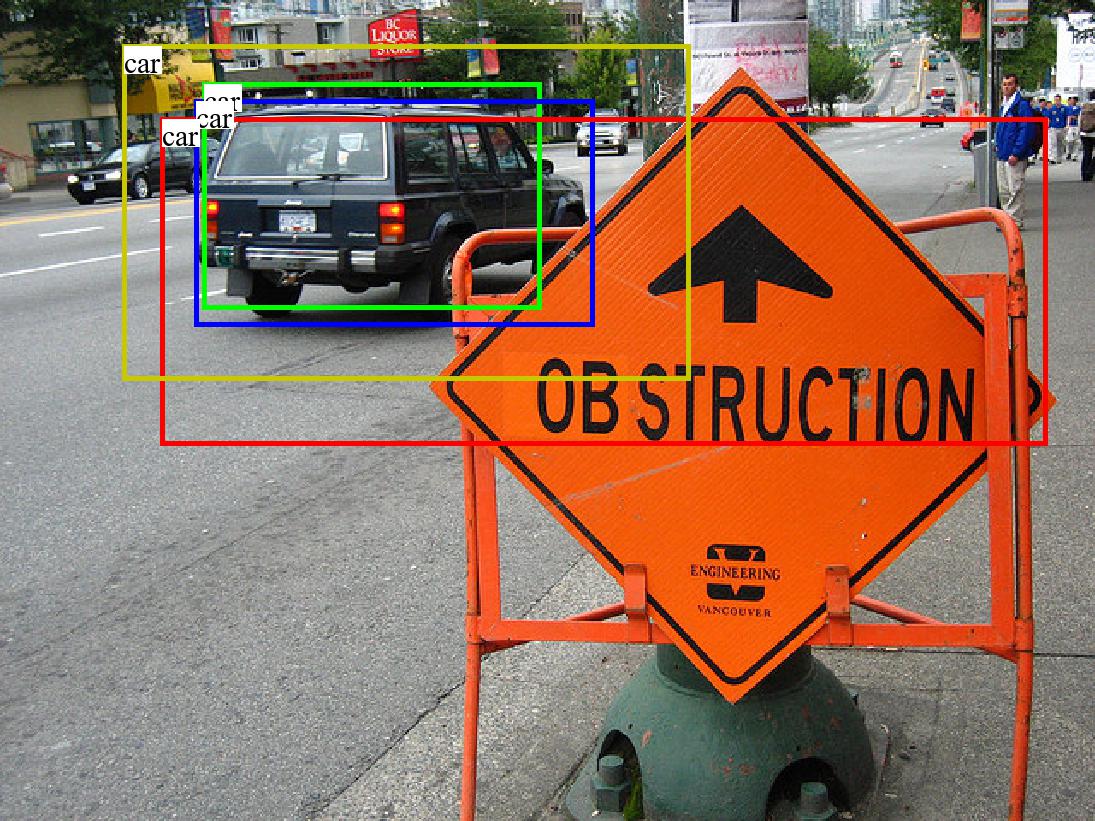}
  \label{fig:sfig6}
\end{subfigure}
\caption{Selected examples of detection results on OccludedCOCO Dataset. Blue box: ground truth; green box: proposals of CA-CompositionalNet via BB Voting; yellow box: proposals of CA-CompositionalNet via RPN; red box: proposals of Faster R-CNN.}
\label{fig:VisualCoco}
\end{figure}

\subsection{Object Detection under Realistic Occlusion}
In the following, we evaluate our model on the OccludedCOCO dataset.
As shown in Table \ref{tab:coco} and Figure \ref{fig:VisualCoco}, our CA-CompositionalNet with robust bounding box voting outperforms Faster R-CNN and CompNet+RPN significantly.
In particular, fully deactivating the context ($\omega=0$) increases the performance compared to the original model ($\omega=0.5$), indicating that too much weight is put on the contextual information in the standard CompNets. 
Furthermore, controlling the prior of the context model to $\omega=0.2$ reaches an optimal performance under strong occlusion where the context is helpful, but does slightly decrease the performance under low occlusion. 

\section{Conclusion}
In this work, we studied the problem of detecting partially occluded objects under occlusion. 
We found that standard deep learning approaches that combine proposal networks with classification networks do not detect partially occluded objects robustly.
Our experimental results demonstrate that this problem has two causes:
1) Proposal networks are more strongly misguided the more context is occupied by the occluders.
2) Classification networks do not classify partially occluded objects robustly.
We made the following contributions to resolve these problems:

\textbf{CompositionalNets for object detection.}
CompositionalNets have proven to classify partially occluded objects robustly. 
We generalize CompositionalNets to object detection by extending their architecture with a detection layer.

\textbf{Robust bounding box voting.} 
We proposed a robust part-based voting mechanism for bounding box estimation by leveraging the unoccluded parts of the object, which enabled the accurate estimation of an object's bounding box even under severe occlusion.

\textbf{Context-aware CompositionalNets.} 
CompositionalNets, and other DCNN-based classifiers, do not separate the representation of the context from that of the object. 
We proposed to segment the object from its context using bounding box annotations and showed how the segmentation can be used to learn a representation in an end-to-end manner that disentangles the context from the object.

\textbf{Acknowledgement.} This work was partially supported by the Swiss National Science Foundation (P2BSP2.181713) and the Office of Naval Research (N00014-18-1-2119).

{\small
\bibliographystyle{ieee_fullname}
\bibliography{egbib}
}

\end{document}